\pdfoutput=1

\documentclass[11pt,twoside]{article}
\usepackage[letterpaper,inner=3.5cm,outer=2.5cm,top=2.5cm,bottom=2.5cm,pdftex]{geometry}
\usepackage[T1]{fontenc}
\usepackage[utf8]{inputenc}
\usepackage{lmodern}
\usepackage[pdftex]{graphicx}
\usepackage[pdftex]{hyperref}
\usepackage{amsfonts}
\usepackage{fancyhdr}
\usepackage{amsmath}
\usepackage{amsthm}
\usepackage{authblk}

\usepackage{mathtools}

\usepackage{amsfonts}       
\usepackage{amsmath}
\usepackage{amsthm}
\usepackage{mathtools}
\usepackage{lmodern}

\newcommand{\mute}[1]{}
\newcommand{\floor}[1]{\left\lfloor #1 \right\rfloor}
\newcommand{\Gset}{\mathbb{G}}
\newcommand{\Rset}{\mathbb{R}}

\newcommand{\Zset}{\mathbb{Z}}

\newcommand{\Xset}{\mathcal{X}}
\newcommand{\Yset}{\mathcal{Y}}
\newcommand{\ZZset}{\mathcal{Z}}

\newcommand*\mpartial{\mathop{}\!\mathrm{d}}
\newcommand*\diff{\mathop{}\!\mathrm{d}}

\newcommand{\Lip}{\mathrm{Lip}}

\newcommand{\norm}[1]{\left\lVert#1\right\rVert}

\newtheorem{theorem}{Theorem}

\newtheorem{corollary}[theorem]{Corollary}

\title{Approximation Capabilities of Neural ODEs\\ and Invertible Residual Networks}

\author{Han Zhang\thanks{e-mail: zhangh4@vcu.edu}}
\author{Xi Gao\thanks{e-mail: gaox2@vcu.edu}}
\author{Jacob Unterman\thanks{e-mail: untermanjw@vcu.edu}}
\author{Tom Arodz\thanks{Corresponding author. e-mail: tarodz@vcu.edu}}
\affil{\mbox{Department of Computer Science}, \mbox{Virginia Commonwealth University} \mbox{Richmond, VA, USA}}
\date{}

\begin{document}
	\sloppy
	\maketitle
	
	\begin{abstract}
Neural ODEs and i-ResNet are recently proposed methods for enforcing invertibility of residual neural models. Having a generic technique for constructing invertible models can open new avenues for advances in learning systems, but so far the question of whether Neural ODEs and i-ResNets can model any continuous invertible function remained unresolved. Here, we show that both of these models are limited in their approximation capabilities. We then prove that any homeomorphism on a $p$-dimensional Euclidean space can be approximated by a Neural ODE operating on a $2p$-dimensional Euclidean space, and a similar result for i-ResNets. We conclude by showing that capping a Neural ODE or an i-ResNet with a single linear layer is sufficient to turn the model into a universal approximator for non-invertible continuous functions.
\end{abstract}

\section{Introduction}

A neural network block is a function $F$ that maps an input vector $x \in \Xset \subset \Rset^p$ to output vector $F(x,\theta) \in  \Rset^{p'}$, and is parameterized by a weight vector $\theta$. We require that $F$ is almost everywhere differentiable  
with respect to both of its arguments, allowing the use of gradient methods for tuning  $\theta$ based on training data and an optimization criterion, and for passing the gradient to preceding network blocks. 

One type of neural building blocks that has received attention in recent years is a {\em residual block} \cite{he2016deep}, where $F(x,\theta)=x+f(x,\theta)$, with $f$ being some differentiable, nonlinear, possibly multi-layer transformation. Input and output dimensionality of a residual block are the same, $p$, and such blocks are usually stacked in a sequence, a ResNet, $x_{t+1}=x_t+f_t(x_t,\theta_t)$. 
Often, the functional form of $f_t$ is the same for all blocks $t \in \left\{1,...T\right\}$ in the sequence. Then, we can represent the sequence through $x_{t+1} - x_t = f_\Theta(x_t,t)$, where $\Theta$ consists of trainable parameters for all blocks in the sequence; the second argument, $t$, allows us to pick the proper subset of parameters, $\theta_t$.
If we allow arbitrary $f_\Theta$, for example a neural network with input and output dimensionality $p$ but with many hidden layers of dimensionality higher than $p$, a sequence of residual blocks can, in principle, model arbitrary mappings $x \rightarrow \phi_T(x)$, where we define $\phi_T(x_0)=x_T$ to be the result of applying the sequence of $T$ residual blocks to the initial input $x_0$.  For example, a linear layer preceded by a deep sequence of  residual blocks is a universal approximator for Lebesgue integrable functions $\Rset^p \rightarrow \Rset$ \cite{lin2018resnet}. 

Recently, models arising from residual blocks have gained attention as a means to construct invertible networks; that is, training a network results in mappings $x_0 \rightarrow x_T$ for which an inverse mapping $x_T \rightarrow x_0$ exist. Ability to train a mapping that is guaranteed to be invertible has practical applications; for example, they give rise to normalizing flows \cite{deco1995nonlinear,rezende2015variational}, which allow for sampling from a complicated, multi-modal probability distribution by generating samples from a simple one, and transforming them through an invertible mapping. For any given architecture for invertible neural networks, it is important to know whether it can be trained to approximate arbitrary invertible mappings, its approximation capabilities are limited. 

\subsection{Invertible Models}
We focus our attention on two invertible neural network architectures: i-ResNet, a constrained ResNet, and  Neural ODE, a continuous generalization of a ResNet. 
\paragraph{Invertible Residual Networks}
While ResNets refer to arbitrary networks with any residual blocks $x_{t+1} = x_t+f_\Theta(x_t,t)$, that is, can have any residual mapping $f_\Theta(x_t,t)$, i-ResNets \cite{behrmann2018invertible}, and their improved variant, Residual Flows \cite{chen2019residual}, are built from  blocks in which $f_\Theta$ is Lipschitz-continuous with constant lower than 1  as a function of $x_t$ for fixed $t$, which we denote by $\Lip(f_\Theta) < 1$. This simple constraint is sufficient \cite{behrmann2018invertible} to guarantee invertibility of the residual network, that is, to make $x_t \rightarrow x_{t+1}$  a one-to-one mapping.

Given the constraint on the Lipschitz constant, an invertible mapping $x \rightarrow 2x$ cannot be performed by a single i-ResNet layer. But a stack of two layers, each of the form $x \rightarrow x+(\sqrt{2}-1)x$ and thus Lipschitz-continuous with constant lower than 1, yields the desired mapping. A single i-ResNet layer $x_{t+1} = (I +f_\Theta)(x_t,t)$, where $I$ is the identity mapping, is $\Lip(I +f_\Theta) = k < 2$, and a composition of $T$ such layers has Lipschitz constant of at most $K=k^T$. Thus, for any finite $K$, it might be possible to approximate any invertible mapping $h$ with $\Lip(h) \leq K$ by a series of i-ResNet layers, with the number of layers depending on $K$. However, the question whether the possibility outlined above true, and i-ResNet have universal approximation capability within the class of invertible continuous mappings, has not been considered thus far. 

\paragraph{Neural Ordinary Differential Equations} Neural ODEs (ODE-Nets) \cite{chen2018neural} are a recently proposed class of differentiable neural network building blocks. 
ODE-Nets were formulated by observing that processing an initial input vector $x_0$ through a sequence of residual blocks can be seen as evolution of $x_t$ in time $t \in \left\{1,...T\right\}$. Then, a residual block (eq. \ref{eq:ResBlock}) 
is a discretization of a continuous-time system of ordinary differential equations (eq. \ref{eq:ODE})
\begin{align}
x_{t+1} - x_t &= f_\Theta(x_t,t), \label{eq:ResBlock} \\
\frac{\mpartial x_t}{\mpartial t} = \lim_{\delta_t \rightarrow 0} \frac{x_{t+\delta_t} - x_t }{\delta_t}   &=f_\Theta(x_t,t). \label{eq:ODE}
\end{align}

The transformation $\phi_T: \Xset \rightarrow \Xset$ taking $x_0$ into $x_T$ realized by an ODE-Net for some chosen, fixed time $T\in \Rset$ is not specified directly through a functional relationship $x \rightarrow f(x)$ for some neural network $f$, but indirectly, through the solutions to the initial value problem (IVP) of the ODE
\begin{equation}
x_T = \phi_T(x_0) = x_0 + \int_0^T f_\Theta(x_t,t) \diff t  \label{eq:IVP}
\end{equation}
involving some underlying neural network $f_\Theta(x_t,t)$ with trainable parameters $\Theta$. By a {\em $p$-ODE-Net} we denote an ODE-Net that takes a $p$-dimensional sample vector on input, and produces a $p$-dimensional vector on output. The underlying network $f_\Theta$ must match those dimensions on its input and output, but in principle can have arbitrary internal architecture, including multiple layers of much higher dimensionality. 

By the properties of ODEs, ODE-Nets are always invertible, we can just reverse the limits of integration, or alternatively integrate  $-f_\Theta(x_t,t)$. The adjoint sensitivity method \cite{pontryagin1962mathematical} based on reverse-time integration of an expanded ODE allows for finding gradients of the IVP solutions $\phi_T(x_0)$ with respect to  parameters $\Theta$ and the initial values $x_0$. This allows training ODE-Nets using gradient descent, as well as combining them with other neural network blocks.
Since their introduction, ODE-Nets have seen improved implementations \cite{rackauckas2019diffeqflux} and enhancements in training and stability \cite{gholami2019anode,zhang2019anodev2}. 

Unlike an unconstrained residual block, a Neural ODE on its own does not have universal approximation capability. Consider a continuous, differentiable, invertible function $f(x)=-x$ on $\Xset=\Rset$. There is no ODE defined on $\Rset$ that would result in $x_T = \phi_T(x_0)=-x_0$. Informally, in ODEs, paths $(x_t,t)$ between the initial value $(x_0,0)$ and final value $(x_T,T)$ have to be continuous and cannot intersect in $\Xset \times \Rset$ for two different initial values, and paths corresponding to $x \rightarrow -x$ and $0 \rightarrow 0$ would need to intersect. By contrast, in an unconstrained residual block sequence, a discrete dynamical system on $\Xset$, we do not have continuous paths, only points at unit-time intervals, with an arbitrary transformation between points; finding an unconstrained ResNet for $x \rightarrow -x$ is easy. While ODE-Nets used out-of-the-box have limited modeling capability, some evidence exists that this limitation can be overcome by changing the way way ODE-Nets are applied. Yet, the question whether they can be turned into universal approximators remains open.

\subsection{Our Contribution}

We analyze the approximation capabilities of ODE-Nets and i-ResNets. The results most closely related to ours have been recently provided by the authors of ANODE \cite{dupont2019augmented}, who focus on a $p$-ODE-Net followed by a linear layer. They provide counterexamples showing that such an architecture is not a universal approximator of $\Rset^p \rightarrow \Rset$ functions. However, they show empirical evidence indicating that expanding the dimensionality and using $q$-ODE-Net for $q>p$ instead of a $p$-ODE-Net has positive impact on training of the model and on its generalization capabilities. The authors of i-ResNet \cite{behrmann2018invertible} also use expanded dimensionality in their experiments, observing that it leads to a modest increase in model's accuracy.

Here, we prove that setting $q=p+1$ is enough to turn Neural ODE followed by a linear layer into a universal approximator for $\Rset^p \rightarrow \Rset$. We show similar result for i-ResNet. Our main focus is on modeling invertible functions -- homeomorphisms -- by exploring pure ODE-Nets and i-ResNets, not capped by a linear layer. We show a class of $\Xset \rightarrow \Xset$ invertible mappings that cannot be expressed by these modeling approaches when they operate within $\Xset$. We then prove that any homeomorphism $\Xset \rightarrow \Xset$, for $\Xset \subset \Rset^p$, can be modeled by a Neural ODE / i-ResNet operating on an Euclidean space of dimensionality $2p$ that embeds $\Xset$ as a linear subspace.

\section{Background on ODEs, Flows, and Embeddings}

This section provides background on invertible mappings and ODEs; we recapitulate standard material, for details see \cite{utz1981embedding,lee2001introduction,brin2002introduction,younes2010shapes}. 

\subsection{Flows }

A mapping $h: \Xset \rightarrow \Xset$ is a {\em homeomorphism} if $h$ is a one-to-one mapping of $\Xset$ onto itself, and both $h$ and its inverse $h^{-1}$ are continuous. Here, we will assume that $\Xset \subset \Rset^p$ for some $p$, and we will use the term $p$-homeomorphism where dimensionality matters.  

A {\em topological transformation group} or a {\em flow} \cite{utz1981embedding} is an ordered triple $(\Xset, \Gset, \Phi)$  involving an additive group $\Gset$ with neutral element 0, and a mapping $\Phi: \Xset \times \Gset \rightarrow \Xset$ such that $\Phi(x,0)=x$  and $\Phi(\Phi(x,s),t)=\Phi(x,s+t)$ for all $x \in \Xset$, all $s,t \in \Gset$. Further, mapping $\Phi(x,t)$ is assumed to be continuous with respect to the first argument.
The mapping $\Phi$ gives rise to a parametric family of homeomorphisms $\phi_t: \Xset \rightarrow \Xset$ defined as $\phi_t(x)=\Phi(x,t)$, with the  inverse being $\phi_t^{-1}=\phi_{-t}$.

Given a flow, an {\em orbit} or a {\em trajectory} associated with $x\in \Xset$ is a subspace $G(x)=\left\{\Phi(x,t): t\in \Gset \right\}$. Given $x,y \in \Xset$, either $G(x)=G(y)$ or $G(x)\cap G(y)=\emptyset$; two orbits are either identical or disjoint, they never intersect. A point $x \in \Xset$ is a {\em fixed point} if $G(x)=\left\{x \right\}$. A {\em path} is a part of the trajectory defined by a specific starting and end points. A path is a subset of $\Xset$; we will also consider a {\em space-time path} composed of points $(x_t,t)$ if we need to make the time evolution explicit. 

A {\em discrete flow} is defined by setting $\Gset=\Zset$. For arbitrary homeomorphism $h$ of $\Xset$ onto itself, we easily get a corresponding discrete flow, an iterated discrete dynamical system, $\phi_0(x)=x$, $\phi_{t+1}=h(\phi_t(x))$, $\phi_{t-1}(x)=h^{-1}(\phi_t(x))$. Setting $f(x)=h(x)-x$ gives us a ResNet $x_{t+1}=x_t+f(x_t)$ corresponding to $h$, though not necessarily an i-ResNet, since there is no $\Lip(f)<1$ constraint.
For Neural ODEs, the type of flow that is relevant is a {\em continuous flow}, defined by setting $\Gset=\Rset$, and adding an assumption that the family of homeomorphisms, the function $\Phi: \Xset \times \Rset \rightarrow \Xset$, is differentiable with respect to its second argument, $t$, with continuous $\diff \Phi / \diff t$. The key difference compared to a discrete flow is that the {\em flow at time $t$}, $\phi_t(x)$, is now defined for arbitrary $t \in \Rset$, not just for integers. We will use the term $p$-flow to indicate that $\Xset \subset \Rset^p$. 

Informally, in a continuous flow the orbits are continuous, and the property that orbits never intersect has consequences for what homeomorphisms $\phi_t$ can result from a flow. Unlike in the discrete case, for a given homeomorphism $h$ there may not be a continuous flow such that $\phi_T=h$ for some $T$. We cannot just set $\phi_T=h$, what is required is a continuous family of homeomorphisms $\phi_t$ such that $\phi_T=h$ and $\phi_0$ is identity -- such family may not exist for some $h$. In such case, a Neural ODE would not be able to model $h$. While i-ResNets are discrete, the way they are constructed may also limit the space of mappings they can model to a subset of all homeomorphisms, even if each residual mapping is made arbitrarily complex within the Lipschitz constraint.

\subsection{Continuous Flows and ODEs}
Given a continuous flow $(\Xset, \Rset, \Phi)$ one can define a corresponding ODE  on $\Xset$ by defining a vector $V(x) \in \Rset^p$ for every $x\in \Xset \subset \Rset^p$ such that $V(x) = \left. \mpartial \Phi(x,t) / \mpartial t \right|_{t=0}$. Then, the ODE $\mpartial x/ \mpartial t =V(x)$
corresponds to continuous flow $(\Xset, \Rset, \Phi)$. Indeed, $\Phi(x_0,T)=x_0 + \int_0^T V(x_t) \diff t$, $\phi_0$ is identity, and $\phi_{(S+T)}(x_0)=\phi_T(\phi_S(x_0))$ for  time-independent $V$. Thus, for any homeomorphism family $\Phi$ defining a continuous flow, there is a corresponding ODE that, integrated for time $T$, models the flow at time $T$, $\phi_T(x)$. 

The vectors of derivatives $V(x) \in \Rset^p$ for all $x \in \Xset$ are continuous over $\Xset$ and are constant in time, and define a {\em continuous vector field} over $\Rset^p$. The ODEs evolving according to such a time-invariant vector field, where the right-hand side of eq. \ref{eq:ODE} depends on $x_t$ but not directly on time $t$, are called {\em autonomous ODEs}, and take the form of $\mpartial x / \mpartial t = f_\Theta(x_t)$. 

Any {\em time-dependent ODE} (eq. \ref{eq:ODE}) can be transformed into an autonomous ODE by removing time $t$ from being a separate argument of $f_\Theta(x_t,t)$, and adding it as part of the vector $x_t$. 
Specifically, we add an additional dimension\footnote{To avoid confusion with $x_t$ indicating time, we use $x[i]$ to denote $i$-th component of vector $x$. } $x[\tau]$ to vector $x$, with $\tau=p+1$. We equate it with time, $x[\tau]=t$, by including $\diff x[\tau] / \diff t = 1$ in the definition of how $f_\Theta$ acts on $x_t$, and including $x_0[\tau] = 0$ in the initial value $x_0$. In defining $f_\Theta$, explicit use of $t$ as a variable is being replaced by using the component $x[\tau]$ of vector $x_t$. The result is an autonomous ODE.

Given time $T$ and an ODE defined by $f_\Theta$,  $\phi_T$, the flow at time $T$, may not be well defined, for example if $f_\Theta$ diverges to infinity along the way.  However, if $f_\Theta$ is well behaved, the flow will exist at least locally around the initial value. 
Specifically, Picard-Lindel{\"o}f theorem states that if an ODE is defined by a Lipschitz-continuous function $f_\Theta(x_t)$, then there exists $\varepsilon > 0$ such that  the flow at time $T$, $\phi_T$, is well-defined and unique for $-\varepsilon < T < \varepsilon$. If exists, $\phi_T$ is a homeomorphism, since the inverse exists and is continuous; simply, $\phi_{-T}$ is the inverse of $\phi_T$. 

\subsection{Flow Embedding Problem for Homeomorphisms}
\label{sec:emb}
Given a $p$-flow, we can always find a corresponding ODE. Given an ODE, under mild conditions, we can find a corresponding flow at time $T$, $\phi_T$, and it necessarily is a homeomorphism. 
Is the class of $p$-flows equivalent to the class of $p$-homeomorphisms, or only to its subset? That is, given a homeomorphism $h$, does a $p$-flow such that $\phi_T=h$ exist? This question is referred to as the problem of embedding the homeomorphism into a flow.

For  a homeomorphism $h: \Xset \rightarrow  \Xset$, its  {\em restricted embedding into a flow}  is a flow $(\Xset, \Rset, \Phi)$ such that $h(x) = \Phi(x,T)$ for some $T$; the flow is restricted to be on the same domain as the homeomorphism. Studies of homeomorphisms on simple domains such as a 1D segment \cite{fort1955embedding} or a 2D plane \cite{andrea1965homeomorphisms} showed that a restricted embedding does not always exist.

An {\em unrestricted embedding into a flow} \cite{utz1981embedding} is a flow $(\Yset, \Rset, \Phi)$ on some space $\Yset$ of dimensionality higher than $p$. It involves a homeomorphism $g: \Xset \rightarrow \ZZset$ that maps $\Xset$ into some subset $\ZZset \subset \Yset$, such that the flow on $\Yset$ results in mappings on $\ZZset$ that are equivalent to $h$ on $\Xset$ for some $T$, that is, $g(h(x)) = \Phi(g(x),T)$. While a solution to the unrestricted embedding problem always exists, it involves a smooth, non-Euclidean manifold $\Yset$.  For a homeomorphism $h: \Xset \rightarrow \Xset$, the manifold $\Yset$, variously referred to as the twisted cylinder \cite{utz1981embedding}, or a suspension under a ceiling function \cite{brin2002introduction}, or a mapping torus \cite{browder1966manifolds}, is a quotient space $\Yset = \Xset \times [0,1] / \sim$ defined through the equivalence relation $(x,1) \sim (h(x),0)$. 
The flow that maps $x$ at $t=0$ to $h(x)$ at $t=1$ and $h(h(x))$ at $t=2$ involves trajectories in $\Xset \times [0,1] / \sim$ in the following way:  for $t$ going from 0 to 1, the trajectory tracks in a straight line from $(x,0)$ to $(x,1)$; in the quotient space $(x,1)$ is equivalent to $(h(x),0)$. Then, for $t$ going from 1 to 2, the trajectory proceeds from $(h(x),0)$ to $(h(x),1) \sim (h(h(x)),0)$.

The fact that the solution to the unrestricted embedding problem involves a flow on a non-Euclidean manifold makes applying it in the context of gradient-trained ODE-Nets difficult.

\section{Approximation of Homeomorphisms  by Neural ODEs}

In exploring the approximation capabilities of Neural ODEs for $p$-homeomorphisms, we will assume that the neural network $f_\Theta(x_t)$ on the right hand side of the ODE is a universal approximator and, if needed, can be made large enough to closely approximate any desired function. Thus, our concern is with what flows can be modeled by a $q$-ODE-Net assuming that $f_\Theta(x_t)$ can have arbitrary internal architecture, including depth and dimensionality, as long as its input-output dimensionality remains fixed at $q$. We consider two scenarios, $q=p$, and $q>p$.

\subsection{Restricting the Dimensionality Limits Capabilities of Neural ODEs}

We show a class of functions that a Neural ODE cannot model, a class that generalizes the $x \rightarrow -x$ one-dimensional example.

\begin{theorem}
	\label{thm:ODEnogoND}
	Let $\Xset=\Rset^p$, and let $\ZZset \subset \Xset$ be a set that partitions $\Xset$ into two or more disjoint, connected subsets $C_i$, for $i=[m]$. Consider a mapping $h: \Xset \rightarrow \Xset$ that
	\begin{itemize}
		\item is an identity transformation on $\ZZset$, that is, $\forall z \in \ZZset, h(z)=z$,
		\item maps some $x \in C_i$ into $h(x) \in C_j$, for $i \neq j$.
	\end{itemize}
	Then, no $p$-ODE-Net can model $h$. 
	\begin{proof}
		A $p$-ODE-Net can model $h$ if a restricted flow embedding of $h$ exists. Suppose that it does, a continuous flow $(\Xset, \Rset, \Phi)$ can be found for $h$ such that the trajectory of $\Phi(x,t)$ is continuous on $t \in [0,T]$ with $\Phi(x,0)=x$ and $\Phi(x,T) = h(x)$ for some $T \in \Rset$, for all $x \in \Xset$. 
		
		If $h$ maps some $x \in C_i$ into $h(x) \in C_j$, for $i \neq j$, the trajectory from $\Phi(x,0) = x \in C_i$ to $\Phi(x,T) = h(x) \in C_j$ crosses $\ZZset$ -- there is $z \in \ZZset$ such that $\Phi(x,\tau)=z$ for some $\tau \in (0,T)$. From uniqueness and reversibility of ODE trajectories, we then have $\Phi(z,-\tau)=x$. From additive property of flows, we have $\Phi(z,T-\tau)=h(x)$. 
		
		Since $h$ is identity over $\ZZset$ and  $\ZZset \subset \Xset$, thus $h(z) = \Phi(z,T) = \Phi(z,0) = z$. That is, the trajectory over time $T$ is a closed curve starting and ending at $z$, and $\Phi(z,t)=\Phi(z,T+t)$ for any $t \in \Rset$. Specifically, $\Phi(z,T-\tau)=\Phi(z,-\tau)=x$. Thus, $h(x)=x$. We arrive at a contradiction with the assumption that $x$ and $h(x)$ are in two disjoint subsets of $\Rset^p$ separated by $\ZZset$. Thus, no $p$-ODE-Net can model $h$.
		
	\end{proof}
\end{theorem}
The result above shows that Neural ODEs applied in the most natural way, with  $q=p$, are severely restricted in the way distinct regions of the input space can be rearranged in order to learn and generalize from the training set, and the restrictions go well beyond requiring invertibility and continuity. 

\subsection{Neural ODEs with Extra Dimensions are Universal Approximators for Homeomorphisms}
\begin{figure*}[t]
	\centering
	\includegraphics[width=0.95\textwidth]{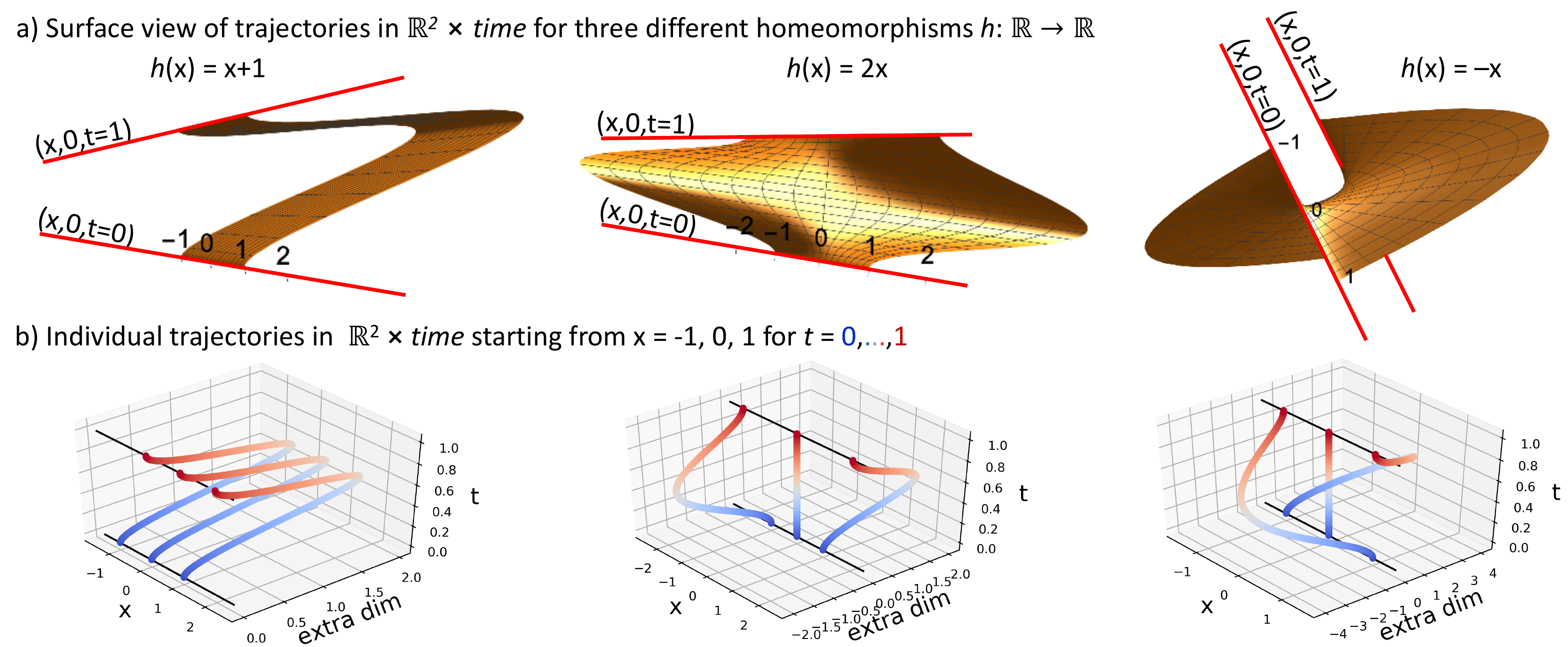} 
	\caption{Trajectories in $\Rset^{2p}$ that embed an $\Rset^p \rightarrow \Rset^p$ homeomorphism, using $f(\tau)=(1-\cos \pi \tau)/2$ and $g(\tau)=(1-\cos 2 \pi \tau)$. Three examples for $p=1$ are shown, including the mapping $h(x)=-x$ that cannot be modeled by Neural ODE on $\Rset^p$, but can in $\Rset^{2p}$.}
	\label{fig:torus}
\end{figure*}

If we allow the Neural ODE to operate on Euclidean space of dimensionality $q>p$, we can approximate arbitrary $p$-homeomorphism $\Xset \rightarrow \Xset$, as long as $q$ is high enough. Here, we show that is suffices to take $q=2p$. 
We construct a mapping from the original problem space, $\Xset \in \Rset^p$ into $\Rset^{2p}$  that \footnote{We use upper subscript $x^{(p)}$ to denote dimensionality of vectors; that is, $0^{(p)}\in \Rset^p$.}
\begin{itemize}
	\item  preserves $\Xset$ as a $p$-dimensional linear subspace consisting of vectors $[x,0^{(p)}]$, 
	\item leads to an ODE that maps $[x,0^{(p)}] \rightarrow  [h(x),0^{(p)}]$.
\end{itemize}
Thus, we provide a solution with a structure that is convenient for out-of-the-box training and inference using Neural ODEs -- it is sufficient to add $p$ zeros to input vectors.   

\begin{theorem}
	\label{thm:main}
	For any homeomorphism $h: \Xset \rightarrow  \Xset$,  $\Xset \subset \Rset^p$,  there exists a $2p$-ODE-Net $\phi_T: \Rset^{2p} \rightarrow \Rset^{2p}$ for $T=1$ such that $\phi_T([x,0^{(p)}]) = [h(x),0^{(p)}]$ for any $x \in \Xset$.
	\begin{proof}
		We prove the existence in a constructive way, by showing a vector field in $\Rset^{2p}$, and thus an ODE, with the desired properties. 
		
		We start with the extended space $(x,\tau)$ with a variable $\tau$ corresponding to time added as the last dimension, as in the construction of an autonomous ODE from time-dependent ODE. We then define a mapping $y(x,\tau): \Rset^p \times \Rset \rightarrow \Rset^{2p}$ that will represent paths starting from $x$ at time $\tau=0$. For $\tau \in [0,1]$, the mapping (see Fig. \ref{fig:torus}) is defined through
		\begin{align}
		y(x,\tau)&=\left[ x + f(\tau)\delta_x, \delta_x g(\tau) \right]. \label{eq:mapping}
		\end{align}
		For each $x$, let $\delta_x \in \Rset^p$ be defined as $\delta_x = h(x)-x$. The functions $f,g: \Rset \rightarrow \Rset$ are required to have continuous first derivative, and have $f(0)=0$, $f(1)=1$, $g(\tau)=0$ iff $\tau \in \Zset$, and the derivatives $\mpartial f / \mpartial \tau$ and $\mpartial g / \mpartial \tau$ are null at $\tau \in \Zset$ and only there. 
		The mapping indeed just adds $p$ dimensions of 0 to $x$ at time $\tau=0$, and at time $\tau=1$ it gives the result of the homeomorphism applied to $x$, again with $p$ dimensions of 0
		\begin{align*}
		y(x,0))&=[x , 0^{(p)}], \\
		y(x,1)&=[x +\delta_x, 0^{(p)}]=[h(x), 0^{(p)}]= y(h(x),0).
		\end{align*}
		For the purpose of constructing an ODE-Net with universal approximation capabilities, $\tau \in [0,1]$ suffices. However, more generally we can define the mapping for $\tau \notin [0,1]$, by setting $y(x,\tau)=y(h^{(\floor{\tau})},\tau - \floor{\tau})$; for example, $y(x,-1.75)=y(h^{-1}(h^{-1}(x)),0.25)$. Intuitively, the mapping $y(x,\tau)$ will provide the position in $\Rset^{2p}$ of the time evolution for duration $\tau$ of an ODE on $\Rset^{2p}$ starting from a position corresponding to $x$. 
		
		For two distinct $x,x' \in \Rset^p$, the paths in $\Rset^{2p}$ given by eq. \ref{eq:mapping} do not intersect at the same position at the same point in time. First, consider the case where $\delta_x$ is not parallel to $\delta_{x'}$. Then, the second set of $p$  variables is equal only if $g(\tau)=0$, that is, only at integer $\tau$. At those time points, the first set of $p$ variables takes iterates of $h$, that is, $...,h^{-1}, x, h(x),h(h(x)),...$, which are different for different $x$. Second, consider $x,x'$ such that $\delta_{x'} = c \delta_x$ for some $c$. Then, either $c \neq 1$ and thus $g(\tau) \neq c g(\tau)$ for all non-integer $\tau$, that is, the second set of $p$ variables are always different except at $\tau$ corresponding to iterates of $h$, which are distinct; or $c=1$, and the second set of $p$ variables are always the same. In the latter case, also $f(\tau)\delta_x = f(\tau)\delta_{x'}$, hence the first set of variables is $x+f(\tau)\delta_x$ is only equal to $x'+f(\tau)\delta_{x'}$ if $x=x'$. Thus, in $\Rset^{2p}$, paths starting from two distinct points do not intersect at the same point in time. Intuitively, we have added enough dimensions to the original space so that we can reroute all trajectories without intersections.  
		
		We have $\tau$ correspond directly to time, that is, $\mpartial \tau / \mpartial t = 1$
		The mapping $y$ has continuous derivative with respect to $t$, defining a vector field over the image of $y$, a subset of $\Rset^{2p}$ 
		\begin{align*}
		\frac{\mpartial y}{\mpartial t}&=\left[ \frac{\mpartial f}{\mpartial t} \delta_x, \frac{\mpartial g}{\mpartial t} \delta_x \right].
		\end{align*}
		From the conditions on $f,g$, we can verify that this time-dependent vector field defined through derivatives of $y(x,\tau)$ with respect to time has the same values for $\tau=0$ and $\tau=1$ for any $x$
		\begin{align*}
		\frac{\mpartial y}{\mpartial t} \left(x,0\right)=[ 0^{(p)}, 0^{(p)}]=\frac{\mpartial y}{\mpartial t} \left(x,1\right)=\frac{\mpartial y}{\mpartial t}  \left(h(x),0\right)
		\end{align*}
		Thus, 
		the vector field is well-behaved at $y(x,1)=y(h(x),0)$, it is continuous over the whole image of $y$. 
		The vector field above is defined over a closed subset $y(x,\tau)$ of $\Rset^{2p}$, and can be (see \cite{lee2001introduction}, Lemma 8.6) extended to the whole $\Rset^{2p}$. A $(2p)$-ODE-Net with a universal approximator network $f_\Theta$ on the right hand side can be designed to approximate the vector field arbitrarily well. The resulting ODE-Net approximates $[x,0^{p}]$ to $[h(x),0^{p}]$.
	\end{proof}
\end{theorem}

Based on the above result, we now have a simple method for training a Neural ODE to approximate a given continuous, invertible mapping $h$ and, for free, obtain also its continuous inverse $h^{-1}$. On input, each sample $x$ is augmented with $p$ zeros. For a given $x$, the output of the ODE-Net is split into two parts. The first $p$ dimensions are connected to a loss function that penalizes deviation from $h(x)$. The remaining $p$ dimensions are connected to a loss function that penalizes for any deviation from 0. Once the network is trained, we can get $h^{-1}$ by using an ODE-Net with $-f_\Theta$ instead of $f_\Theta$ used in the trained ODE-Net.

\section{Approximation of Homeomorphisms  by i-ResNets}

\subsection{Restricting the Dimensionality Limits Capabilities of i-ResNets}

We show that similarly to Neural ODEs, i-ResNets cannot model a simple $f(x)\rightarrow -x$ homeomorphism $\Rset \rightarrow \Rset$, indicating that their approximation capabilities are limited. 

\begin{theorem}
	\label{thm:iResNetMinusX}
	Let $F_n(x) = (I+f_n)\circ (I+f_{n-1}) \circ \cdots \circ (I+f_{1}) (x)$ be an $n$-layer i-ResNet, and let $x_0 = x$ and $x_n=F_n(x_0)$. If $\Lip(f_i) < 1$ for all $i = 1,...,n$, then there are is no number $n \geq 1$ and no functions $f_i$ for all $i = 1,...,n$ such that $x_n = -x_0$. 
	\begin{proof}
		Consider $a_0 \in \Rset$ and $b_0=a_0+\delta_0$. Then, $a_1 = a_0 + f_1(a_0)$ and $b_1=a_0+\delta_0 + f_1(a_0+\delta_0)$. From $\Lip(f_1) < 1$ we have that $|f_1(a_0+\delta_0)- f_1(a_0)| < |\delta_0|$. Let $\delta_1 =b_1 - a_1$. Then, we have
		\begin{align*} 
		\delta_1 &= a_0 + \delta_0 + f_1(a_0+\delta_0) - a_0 - f_1(a_0) \\
		& = \delta_0 + f_1(a_0+\delta_0) - f_1(a_0), \\
		\delta_1 & > \delta_0 - |\delta_0|, \\
		\delta_1 & < \delta_0 + |\delta_0|.
		\end{align*}
		That is, $\delta_1$ has the same sign as $\delta_0$. Thus, applying the reasoning to arbitrary $i$, $i+1$ instead of $0,1$, if $a_i < b_i$, then $a_{i+1} < b_{i+1}$, and  if $a_i > b_i$, then $a_{i+1} > b_{i+1}$, for all $i= 0,...,n-1$. Assume we can construct an i-ResNet $F_n$ such that $F_n(0)=0$; then $F_n(x) > 0$ for any $x>0$, and $F_n(x)$ cannot map $x$ into $-x$.   
	\end{proof}
\end{theorem}

The result above leads to a more general observation about paths in spaces of dimensionality higher than one. As with ODE-Nets, we will use $p$-i-ResNet to denote an i-ResNet operating on $\Rset^p$. 
\begin{corollary}
	\label{col:ResNetPaths}
	Let the straight line connecting $x_t \in \Rset^p$ to $x_{t+1}=x_t+f(x_t) \in \Rset^p$ be called an {\em extended path} $x_t \rightarrow x_{t+1}$ of a time-discrete topological transformation group on $\Xset \in \Rset^p$. In $p$-i-ResNet, for $x_t \neq x'_t$, extended paths  $x_t \rightarrow x_{t+1}$ and $x'_t \rightarrow x'_{t+1}=x'_{t}$ do not intersect.  
	\begin{proof}
		For two extended paths to intersect, vectors  $x_t,x'_t,  x_{t+1}, x'_{t+1}$ have to be co-planar. If we restrict attention to dynamics starting form $x_t,x'_t$, we can view it as a one-dimensional system, with the space axis parallel to $x_t -- x'_t$, and time axis orthogonal to it. 
		
		First, consider the situation where the line connecting $x_t,x'_t$ is  parallel, in the original $\Rset^p$ space, to the line connecting $x_{t+1}, x'_{t+1}$. Applying Theorem \ref{thm:iResNetMinusX} shows that if $x'_t$ is above $x_t$, then  $x'_{t+1}$ is above $x_{t+1}$; extended paths do not intersect in this case. 
		
		In the more general case, without loss of generality assume that $x'_{t+1}$ is the one extending farther from the $x_t -- x'_t$ line than $x_{t+1}$. We can construct another i-ResNet, keeping  $x_t \rightarrow x_t + f(x_t)$ but with $x'_t \rightarrow x'_t + cf(x'_t)$, for $c<1$ such that we arrive at the parallel case above; $\Lip(cf)<1$. The intersection of extended paths for both i-ResNets, if exists, is at the same position. But the second i-ResNet is the parallel case analyzed above, with no intersection. 

	\end{proof}
\end{corollary}

The result allows us to show that i-ResNets faces a similar constraint in its capabilities as Neural ODEs
\begin{theorem}
	Let $\Xset=\Rset^p$, and let $\ZZset \subset \Xset$ and $h: \Xset \rightarrow \Xset$ be the same as in Theorem \ref{thm:ODEnogoND}. Then, no $p$-i-ResNet can model $h$. 
	\begin{proof}
		Consider a $T$-layered i-ResNet on $\Xset$, giving riving rise to extended space-time paths in $\Xset \times [0,T]$, with integer $t \in [0,T]$ corresponding to activations in subsequent layers. For any $x \in \ZZset$, the extended path in $\Xset \times [0,T]$ starts at $(x,0)$ and ends at $(x,T)$. Since i-ResNet layers are continuous transformations, the union of all extended paths arising from $\ZZset$ is a simply connected subset of $\Xset \times [0,T]$; it has no holes and partitions $\Xset \times [0,T]$ into separate regions. Since extended paths cannot intersect, $(x,T)$ remains in the same region as $(x,0)$, which is in contradiction with mapping $h$. 		
	\end{proof}
\end{theorem}
The proof shows that the limitation in capabilities of the two architectures for invertible mappings analyzed here arises from the fact that paths in invertible mappings constructed through NeuralODEs and i-ResNets are not allowed to intersect and from continuity in $\Xset$. 

\subsection{i-ResNets with Extra Dimensions are Universal Approximators for Homeomorphisms}

Similarly to Neural ODEs, expanding the dimensionality of the i-ResNet from $p$ to $2p$ by adding zeros on input guarantees that any $p$-homeomorphism can be approximated, as long as its Lipschitz constant is finite and an upper bound on it is known during i-ResNet architecture construction.

\begin{theorem}
	\label{thm:mainResNet}
	For any homeomorphism $h: \Xset \rightarrow  \Xset$,  $\Xset \subset \Rset^p$ with $\Lip(h) \leq k$,  there exists a $2p$-i-ResNet $\phi: \Rset^{2p} \rightarrow \Rset^{2p}$ with $\lfloor k+4 \rfloor$ residual layers such that $\phi([x,0^{(p)}]) = [h(x),0^{(p)}]$ for any $x \in \Xset$.
	\begin{proof}
		For a given invertible i-ResNet approximating $h$, define a possibly non-invertible mapping $\delta(x) = (h(x)-x)/T$, where $T=\lfloor k+1 \rfloor$; we have $\Lip(\delta(x)) < 1$.  An i-ResNet that models $h$ using $T+3$ layers $\phi_i$ for $i=0,...,T+2$ can be constructed in the following way:
		\begin{align*}
		\phi_0([x,0]) &\rightarrow [x,0]+[0, \delta(x) ],  \\
		\phi_{i}([z,y]) &\rightarrow [z,y]+[yT/(T+1), 0] \;\; i=1,...,T+1, \\
		\phi_{T+2}([h(x),\delta(x) ]) &\rightarrow [h(x),\delta(x) ]+[0, -\delta(x)]&  \\
		\end{align*}	
		The first layer maps $x$ into $\delta(x)$ and stores it in the second set of $p$ activations. The subsequent $T+1$ layers progress in a straight line from $[x,\delta(x)]$ to $[h(x),\delta(x)]$ in $T+1$ constant-length steps, and the last layer restores null values in the second set of $p$ activations. 
		
		All layers are continuous mappings.
		The residual part of the first layer has Lipschitz constant below one, since $\Lip(\delta(x)) < 1$. 
		The middle layers have residual part constant in $z$ and contractive in $y$, with Lipschitz constant below one. 
		The residual part of the last layer is a mapping of the form $[h,\delta] \rightarrow [0,-\delta]$.
		For a pair $x,x' \in \Xset$, let $h=h(x), h'=h(x')$, $\delta=\delta(x), \delta'=\delta(x')$. We have $\norm{[0,-\delta]-[0,-\delta']} = \norm{\delta-\delta'} \leq \norm{[h,\delta]-[h',\delta']}$, with equality only if $h=h'$. From invertibility of $h(x)$ we have that $h=h'$ implies $x=x'$ and thus $\delta=\delta'$; hence, the residual part of the last layer also has Lipschitz constant below one.
	\end{proof}
\end{theorem}

The theoretical construction above suggests that while on the order of $k$ layers may be needed to approximate arbitrary homeomorphism $h(x)$ with $\Lip(h) \leq k$, only the first and last layers depend on $h(x)$ and need to be trained; the middle layers are simple, fixed linear layers. The last layer for $x \rightarrow h(x)$ is the same as the first layer of $h(x) \rightarrow x$ would be, but the inverse of the first layer, but since i-ResNet construct invertible mappings  $x \rightarrow x + f(x)$ using possibly non-invertible $f(x)$, it has to be trained along with the first layer. 

The construction for i-ResNets is similar to that for NeuralODEs, except one does not need to enforce differentiability in the time domain, hence we do not need smooth accumulation and removal of $\delta(x)$ in the second set of $p$ activations, and the movement from $x$ to $h(x)$ in the original $p$ dimensions does not need to be smooth. In both cases, the transition from $x$ to $h(x)$ progresses along a straight line in the first $p$ dimensions, with the direction of movement stored in the second set of $p$ variables. 

\section{Invertible Networks capped by a Linear Layer are Universal Approximators}

We show that a Neural ODE or an i-ResNet followed by a single linear layer can approximate functions, including non-invertible functions, equally well as any traditional feed-forward neural network. Since networks with shallow-but-wide fully-connected architecture \cite{cybenko1989approximation,hornik1991approximation}, or narrow-but-deep unconstrained ResNet-based architecture \cite{lin2018resnet} are universal approximators, so are ODE-Nets and i-ResNets. Consider a function $\Rset^p \rightarrow \Rset^r$. For any $(x,y)$ such that $y=f(x)$, the mapping $[x,0]\rightarrow [x,y]$ is a $(p+r)$-homeomorphism, and as we have shown, can be approximated by a $2(p+r)$-ODE-Net or $2(p+r)$-i-ResNet; $y$ can be extracted from the result by a simple linear layer. Through a simple construction, we show that actually using just $p+r$ dimensions is sufficient. 

\begin{theorem}
	Consider a neural network $F: \Rset^p \rightarrow \Rset^r$ that approximates function $f: \Xset \rightarrow \Rset^r$ that is Lebesgue integrable for each of the $r$ output dimensions, with $\Xset \subset \Rset^p$ being a compact subset. For $q= p+r$, there exists a linear layer-capped $q$-ODE-Net that can perform the mapping $F$. If $f$ is Lipschitz, there also is a linear layer-capped $q$-i-ResNet for $F$. 
	\begin{proof}
		Let $G$ be a neural network that takes input vectors $x^{(q)}=[x^{(p)},x^{(r)}]$ and produces $q$-dimensional output vectors $y^{(q)}=[y^{(p)},y^{(r)}]$, where $y^{(r)}=F(x^{(p)})$ is the desired transformation. $G$ is constructed as follows: use $F$ to produce $y^{(r)}=F(x^{(p)})$, ignore $x^{(r)}$, and always output $y^{(p)}=0$. 
		Consider a $q$-ODE-Net defined through $\mpartial x / \mpartial t = G(x_t)=[0^{(p)},F(x_t^{(p)})]$. Let the initial value be $x_0 = [x^{(p)},0^{(r)}]$. The ODE will not alter the first $p$ dimensions throughout time, hence for any $t$, $F(x_t^{(p)})= y^{(r)}$. After time $T=1$, we will have
		\begin{align*}
		x_T &= x_0 + \int_0^1 G(x_t) \diff t = [x^{(p)},0^{(r)}] + \int_0^1 [0^{(p)},y^{(r)}] \diff t 
		\\&= [x^{(p)},F(x^{(p)})].
		\end{align*}
		Thus, for any $x \in \Rset^p$, the output $F(x)$ can be recovered from the output of the ODE-Net by a simple, sparse linear layer that ignores all dimensions except the last $r$, which it returns. A similar construction can be used for defining layers of i-ResNet. We define $k$ residual layers, each with residual mapping $[x^{(p)},...] \rightarrow [x^{(p)},...]  + [0^{(p)},F(x^{(p)})/k]$. If $\Lip(F) < k$, then  the residual mapping $[x^{(p)},...] \rightarrow [0^{(p)},F(x^{(p)})/k]$ has Lipschitz constant below 1.
	\end{proof}
\end{theorem}

\section{Experimental Results}

\subsection{i-ResNets}

We tested whether i-ResNet operating in one dimension can learn to perform the $x \rightarrow -x$ mapping, and whether adding one more dimension has impact on the ability learn the mapping. To this end, we constructed a network with five residual blocks. In each block, the residual mapping is a single linear transformation, that is, the residual block is $x_{t+1}=x_t + Wx_t$. We used the official i-ResNet PyTorch package \cite{behrmann2018invertible} that relies on spectral normalization \cite{miyato2018spectral} to limit the Lipschitz constant to less than unity. We trained the network on a set of 10,000 randomly generated values of $x$ uniformly distributed in $[-10,10]$ for 100 epochs, and used an independent test set of 2,000 samples generated similarly. 

For the one-dimensional $x \rightarrow -x$ and the two-dimensional $[x,0] \rightarrow [-x,0]$ target mapping, we used MSE as the loss. Adding one extra dimension results in successful learning of the mapping, confirming Theorem \ref{thm:mainResNet}. The test MSE on each output is below $10^{-10}$; the network learned to negate $x$, and to bring the additional dimension back to null, allowing for invertibility of the model. For the i-ResNet operating in the original, one-dimensional space, learning is not successful (MSE of 33.39), the network learned a mapping $x \rightarrow cx$ for a small positive $c$, that is, the mapping closest to negation of $x$ that can be achieved while keeping non-intersecting paths, confirming experimentally Corollary \ref{col:ResNetPaths}. 

\begin{figure*}[t]
	\centering
	\includegraphics[width=0.3\textwidth]{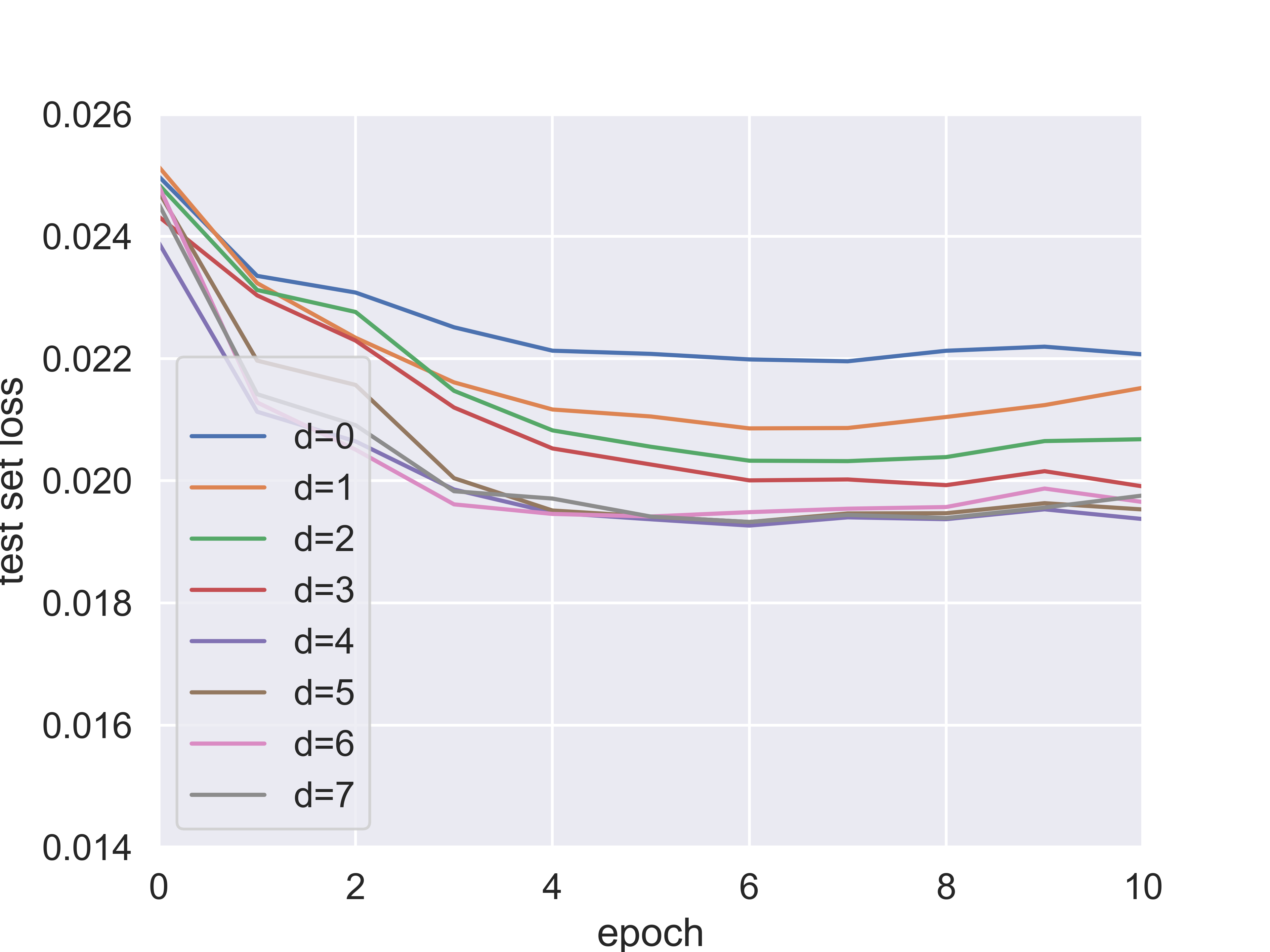} 
	\includegraphics[width=0.3\textwidth]{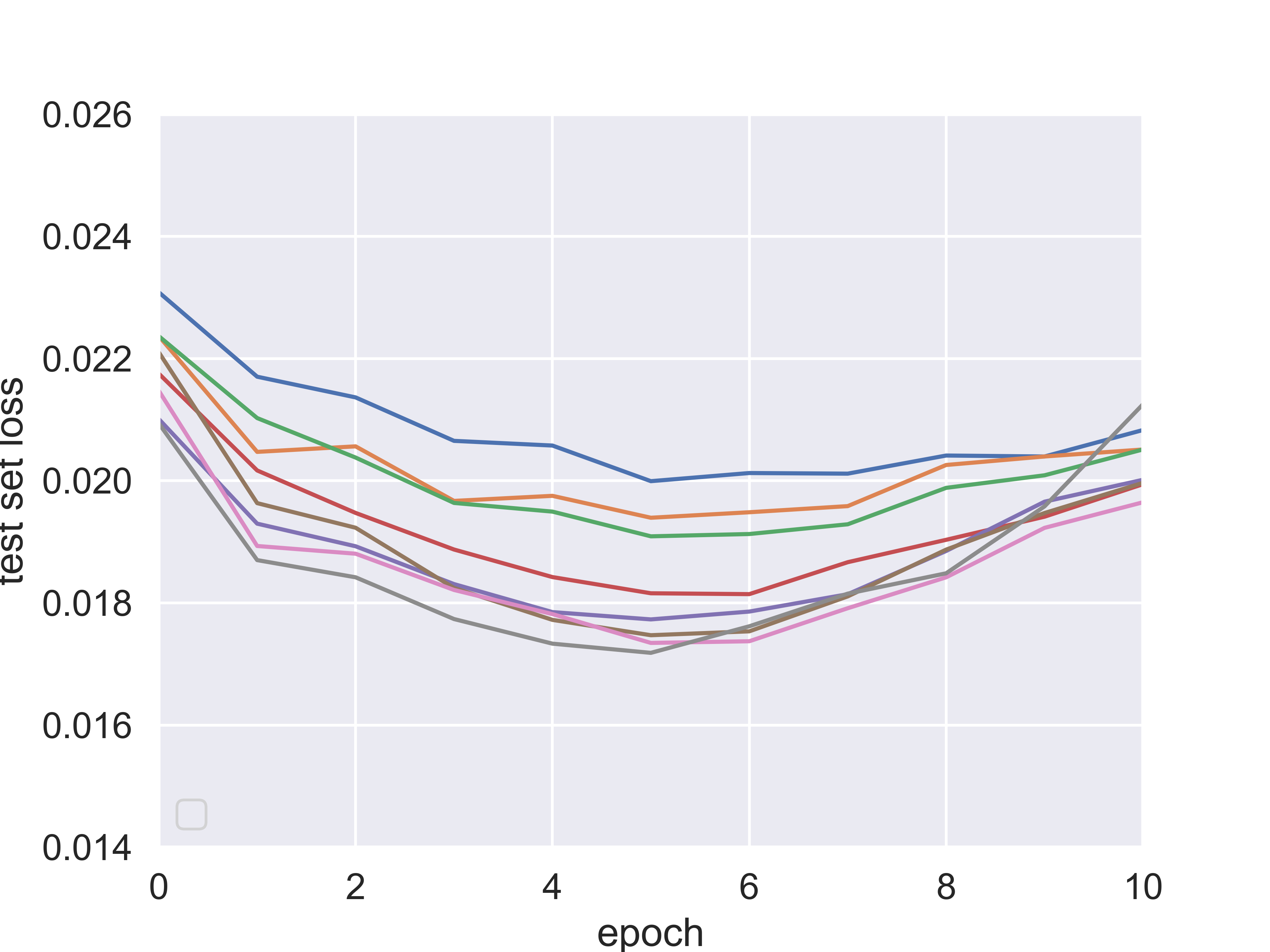} 
	\includegraphics[width=0.3\textwidth]{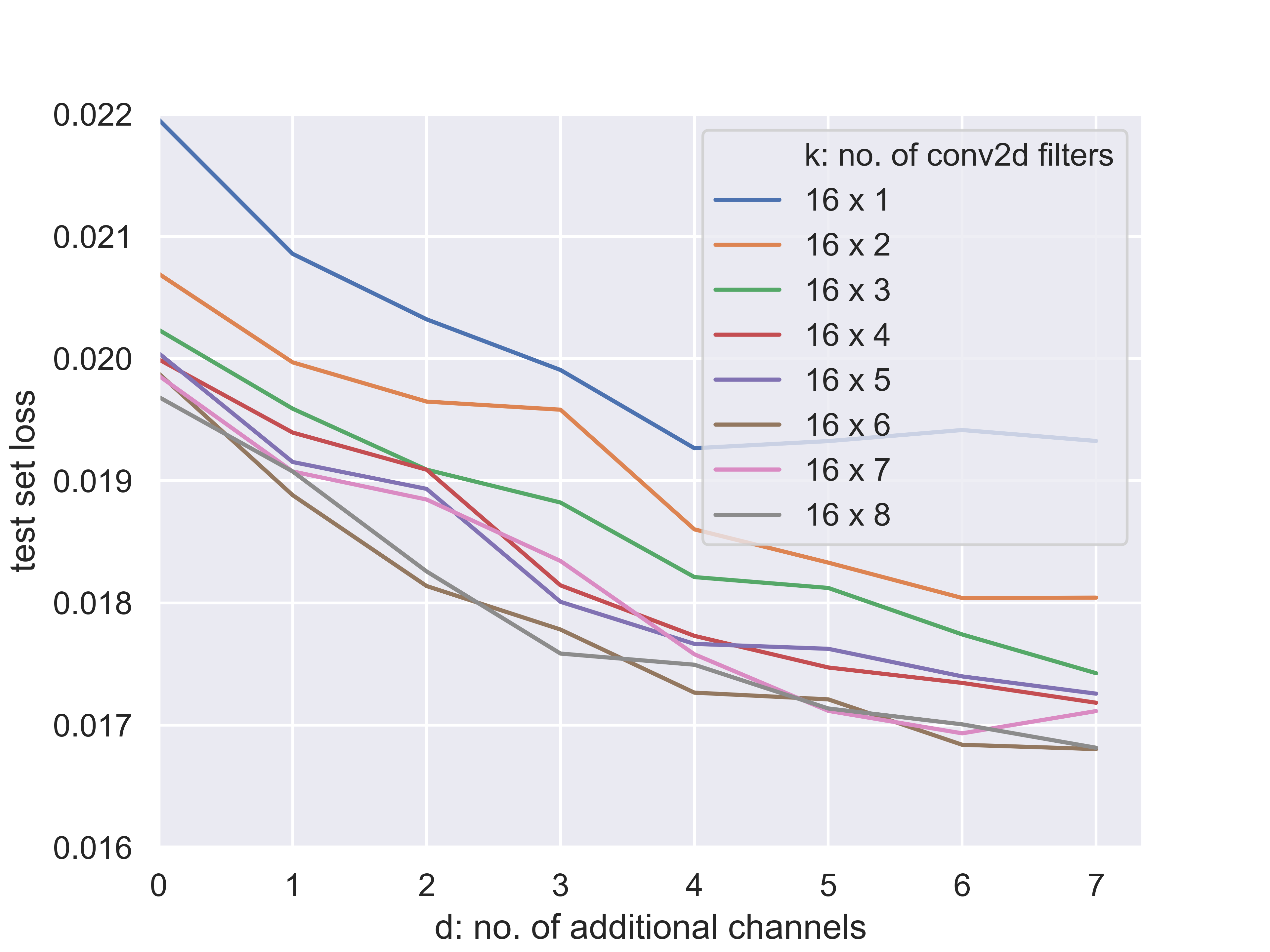} 
	\caption{{\bf Left and center:} test set cross-entropy loss, for increasing number $d$ of null channels added to RGB images. For each $d$, the input images have dimensionality $32\times 32\times (3+d)$. Left: ODE-Net with $k$=64 convolutional filters; center: $k$=128 filters. {\bf Right:} Minimum of test set cross-entropy loss across all epochs as a function of $d$, the number of null channels added to input images, for ODE-Nets with different number of convolutional filters, $k$.}
	\label{fig:teL} 
\end{figure*}

\subsection{Neural ODEs}

We performed experiments to validate if the $q\geq 2p$-dimensions threshold beyond which any $p$-homeomorphism can be approximated by a $q$-ODE-Net can be observed empirically in a practical classification problem. We used the CIFAR10 dataset \cite{krizhevsky2009learning} that consists of 32 x 32 RGB images, that is, each input image has dimensionality of $p=32\times 32\times 3$. We constructed a series of $q$-ODE-Nets with dimensionality $q\geq p$, and for each measured the cross-entropy loss for the problem of classifying CIFAR10 images into one of ten classes. We used the default split of the dataset into 50,000 training and 10,000 test images. 
In designing the architecture of the neural network underlying the ODE we followed ANODE \cite{dupont2019augmented}. Briefly, the network is composed of three 2D convolutional layers. The first two convolutional layers use $k$ filters, and the last one uses the number of input channels as the number of filters, to ensure that the dimensionalities of the input and output of the network match. The convolution stack is followed by a ReLU activation function. A linear layer, with softmax activation and cross-entropy loss, operates on top the ODE block. We used torchdiffeq package \cite{chen2018neural} and trained on a single NVIDIA Tesla V100 GPU card.

To extended the dimensionality of the space in which the ODE operates, we introduce additional null channels on input, that is, we use input images of the form $32\times 32\times (3+d)$. Then, to achieve $q=2p$, we need $d=3$. We tested $d \in \left\{  0,...,7\right\}$.  To analyze how the capacity of the network interplays with the increases in input dimensionality, we also experimented with varying the number of convolutional filters, $k$, in the layers inside the ODE block. 

The results in Fig. \ref{fig:teL} show that once the networks with small network capacity, below 64 filters, behaves differently than networks with 64 or more filters. Once the network capacity is high enough, at 64 filters or more, adding dimensions past beyond 3, that is, beyond $2p$, results in slower decrease in test set loss. To quantify if this slowdown is likely to arise by chance, we calculated the change in test set loss $\ell_d$ for dimensionality $d$ as the $d$ increases by one, $\delta_d=\ell_d-\ell_{d-1}$, for $d$=1,...,7. We pooled the results from experiments with 64 convolution filter or more. Two-tailed nonparametric Mann-Whitney U test between $\delta_1,...,\delta_3$  and $\delta_4,...,\delta_7$ shows the change of trend is significant ($p$=.002).

\section*{Acknowledgments}
T.A. is supported by NSF grant IIS-1453658.

\bibliographystyle{alpha}

\begin{thebibliography}{MKKY18}
	
	\bibitem[And65]{andrea1965homeomorphisms}
	Stephen~A Andrea.
	\newblock On homeomorphisms of the plane, and their embedding in flows.
	\newblock {\em Bulletin of the American Mathematical Society}, 71(2):381--383,
	1965.
	
	\bibitem[BGC{\etalchar{+}}19]{behrmann2018invertible}
	Jens Behrmann, Will Grathwohl, Ricky~TQ Chen, David Duvenaud, and
	J{\"o}rn-Henrik Jacobsen.
	\newblock Invertible residual networks.
	\newblock In {\em International Conference on Machine Learning}, pages
	573--582, 2019.
	
	\bibitem[Bro66]{browder1966manifolds}
	William Browder.
	\newblock Manifolds with $\pi _1 = z$.
	\newblock {\em Bulletin of the American Mathematical Society}, 72(2):238--244,
	1966.
	
	\bibitem[BS02]{brin2002introduction}
	Michael Brin and Garrett Stuck.
	\newblock {\em Introduction to dynamical systems}.
	\newblock Cambridge university press, 2002.
	
	\bibitem[CBDJ19]{chen2019residual}
	Tian~Qi Chen, Jens Behrmann, David~K Duvenaud, and J{\"o}rn-Henrik Jacobsen.
	\newblock Residual flows for invertible generative modeling.
	\newblock In {\em Advances in Neural Information Processing Systems}, pages
	9913--9923, 2019.
	
	\bibitem[CRBD18]{chen2018neural}
	Tian~Qi Chen, Yulia Rubanova, Jesse Bettencourt, and David~K Duvenaud.
	\newblock Neural ordinary differential equations.
	\newblock In {\em Advances in Neural Information Processing Systems}, pages
	6571--6583, 2018.
	
	\bibitem[Cyb89]{cybenko1989approximation}
	George Cybenko.
	\newblock Approximation by superpositions of a sigmoidal function.
	\newblock {\em Mathematics of control, signals and systems}, 2(4):303--314,
	1989.
	
	\bibitem[DB95]{deco1995nonlinear}
	Gustavo Deco and Wilfried Brauer.
	\newblock Nonlinear higher-order statistical decorrelation by volume-conserving
	neural architectures.
	\newblock {\em Neural Networks}, 8(4):525--535, 1995.
	
	\bibitem[DDT19]{dupont2019augmented}
	Emilien Dupont, Arnaud Doucet, and Yee~Whye Teh.
	\newblock Augmented neural {ODEs}.
	\newblock {\em arXiv preprint arXiv:1904.01681}, 2019.
	
	\bibitem[For55]{fort1955embedding}
	Marion~Kirkland Fort.
	\newblock The embedding of homeomorphisms in flows.
	\newblock {\em Proceedings of the American Mathematical Society},
	6(6):960--967, 1955.
	
	\bibitem[GKB19]{gholami2019anode}
	Amir Gholami, Kurt Keutzer, and George Biros.
	\newblock {ANODE:} unconditionally accurate memory-efficient gradients for
	neural {ODEs}.
	\newblock {\em arXiv preprint arXiv:1902.10298}, 2019.
	
	\bibitem[Hor91]{hornik1991approximation}
	Kurt Hornik.
	\newblock Approximation capabilities of multilayer feedforward networks.
	\newblock {\em Neural networks}, 4(2):251--257, 1991.
	
	\bibitem[HZRS16]{he2016deep}
	Kaiming He, Xiangyu Zhang, Shaoqing Ren, and Jian Sun.
	\newblock Deep residual learning for image recognition.
	\newblock In {\em Proceedings of the IEEE conference on Computer Vision and
		Pattern Recognition}, pages 770--778, 2016.
	
	\bibitem[Kri09]{krizhevsky2009learning}
	Alex Krizhevsky.
	\newblock Learning multiple layers of features from tiny images.
	\newblock 2009.
	
	\bibitem[Lee01]{lee2001introduction}
	John~M Lee.
	\newblock {\em Introduction to smooth manifolds}.
	\newblock Springer, 2001.
	
	\bibitem[LJ18]{lin2018resnet}
	Hongzhou Lin and Stefanie Jegelka.
	\newblock {ResNet} with one-neuron hidden layers is a universal approximator.
	\newblock In {\em Advances in Neural Information Processing Systems}, pages
	6169--6178, 2018.
	
	\bibitem[MKKY18]{miyato2018spectral}
	Takeru Miyato, Toshiki Kataoka, Masanori Koyama, and Yuichi Yoshida.
	\newblock Spectral normalization for generative adversarial networks.
	\newblock In {\em International Conference on Learning Representations}, 2018.
	
	\bibitem[PMBG62]{pontryagin1962mathematical}
	Lev~Semenovich Pontryagin, EF~Mishchenko, VG~Boltyanskii, and RV~Gamkrelidze.
	\newblock The mathematical theory of optimal processes.
	\newblock 1962.
	
	\bibitem[RIM{\etalchar{+}}19]{rackauckas2019diffeqflux}
	Chris Rackauckas, Mike Innes, Yingbo Ma, Jesse Bettencourt, Lyndon White, and
	Vaibhav Dixit.
	\newblock {DiffEqFlux.jl}-a {Julia} library for neural differential equations.
	\newblock {\em arXiv preprint arXiv:1902.02376}, 2019.
	
	\bibitem[RM15]{rezende2015variational}
	Danilo Rezende and Shakir Mohamed.
	\newblock Variational inference with normalizing flows.
	\newblock In {\em International Conference on Machine Learning}, pages
	1530--1538, 2015.
	
	\bibitem[Utz81]{utz1981embedding}
	WR~Utz.
	\newblock The embedding of homeomorphisms in continuous flows.
	\newblock In {\em Topology Proc}, volume~6, pages 159--177, 1981.
	
	\bibitem[Whi44]{whitney1944singularities}
	Hassler Whitney.
	\newblock The singularities of a smooth $n$-manifold in (2$n$- 1)-space.
	\newblock {\em Ann. of Math}, 45(2):247--293, 1944.
	
	\bibitem[You10]{younes2010shapes}
	Laurent Younes.
	\newblock {\em Shapes and diffeomorphisms}, volume 171.
	\newblock Springer, 2010.
	
	\bibitem[ZYG{\etalchar{+}}19]{zhang2019anodev2}
	Tianjun Zhang, Zhewei Yao, Amir Gholami, Kurt Keutzer, Joseph Gonzalez, George
	Biros, and Michael Mahoney.
	\newblock {ANODEV2}: A coupled neural {ODE} evolution framework.
	\newblock {\em arXiv preprint arXiv:1906.04596}, 2019.
	
\end{thebibliography}
\newcommand{\etalchar}[1]{$^{#1}$}
\newcommand{\noopsort}[1]{}

\end{document}